\documentclass[runningheads]{llncs}
\usepackage[T1]{fontenc}
\usepackage{graphicx}
\usepackage{url}
\usepackage{hyperref}
\usepackage{framed,multirow}
\usepackage{makecell}
\usepackage{subfigure}
\usepackage{color}

\begin{document}
\title{SGDraw: Scene Graph Drawing Interface using Object-Oriented Representation}
\titlerunning{SGDraw}
% \author{Tianyu Zhang\inst{1}\orcidID{0009-0007-7096-2802} \and
% Xusheng Du\inst{1} \orcidID{0009-0008-1086-2081}\and
% Chia-Ming Chang\inst{2} \orcidID{0000-0003-0390-6361}\and
% Xi Yang\inst{3} \orcidID{0000-0001-5039-3680}\and
% Haoran Xie\inst{1}\orcidID{0000-0002-6926-3082}}

\author{Tianyu Zhang\inst{1}\and
Xusheng Du\inst{1}\and
Chia-Ming Chang\inst{2}\and
Xi Yang\inst{3}\and
Haoran Xie\inst{1}}
\authorrunning{T. Zhang et al.}
\institute{Japan Advanced Institute of Science and Technology, Ishikawa, Japan \and
The University of Tokyo, Tokyo, Japan\and
Jilin University, Jilin, China}
\maketitle              % typeset the header of the contribution
\begin{abstract}
Scene understanding is an essential and challenging task in computer vision. To provide the visually-
grounded graphical structure of an image, the scene graph has received increased attention due to offering 
explicit grounding of visual concepts. Previous works commonly get scene graphs by using ground-truth 
annotations or generating from the target images. However, drawing a proper scene graph for image 
retrieval, image generation, and multi-modal applications is difficult. The conventional scene graph 
annotation interface is not easy to use and hard to revise the results. The automatic scene graph 
generation methods using deep neural networks only focus on the objects and relationships while 
disregarding attributes. In this work, we propose SGDraw, a scene graph drawing interface that uses object-
oriented representation to help users interactively draw and edit scene graphs. SGDraw provides a web-based 
scene graph annotation and creation tool for scene understanding applications. To verify the effectiveness 
of the proposed interface, we conducted a comparison study with the conventional tool and the user 
experience study.  The results show that SGDraw can help create scene graphs with richer details and 
describe the images more accurately than traditional bounding box annotations. We believe the proposed 
SGDraw can be useful in various vision tasks, such as image generation and retrieval. The project source 
code is available at \href{https://github.com/zty0304/SGDraw}{https://github.com/zty0304/SGDraw}.
\end{abstract}
\section{Introduction}
Scene graph is a common and popular way to describe scene understanding, first proposed for image retrieval tasks to search for images with similar descriptions in image datasets~\cite{johnson2015image}. In addition, scene graphs are used for a wide range of vision applications, such as image retrieval~\cite{johnson2015image,qi2017online}, image captioning~\cite{chen2020say}, visual reasoning~\cite{shi2019explainable}, visual answering~\cite{hildebrandt2020scene}, and robotics~\cite{amiri2022reasoning}. The conventional approaches for scene graph creation have been mainly based on a manually annotated image detection dataset, and the annotation tasks are usually time-consuming and laborious. The visual genome dataset~\cite{krishna2017visual} is the most common and popular dataset for scene graphs, with around $10^5$ annotated images. However, the annotated scene graphs have challenging issues, such as sparse image annotation, repeated labels, and unreasonable annotation. In addition, these scene graphs are difficult to be modified and produced in conventional computer vision tasks. 

To solve these issues, one possible solution is the automatic generation of scene graph generation with deep learning approaches. The task of scene graph generation is to generate a corresponding graph-structure representation from an image, and abstract the objects and object relationships from the images. The automatic generation approaches can only generate limited scene graphs on commonly used object categories and predicates, and have difficulty for rare and previously unseen compositions~\cite{knyazev2021generative,knyazev2020graph}. In addition, these automatic approaches can complete the generation of objects and relationships but lack the description of attributes, which makes the generated scene graph incomplete. Therefore, we aim to provide a novel scene graph drawing interface with users in the loop.

\begin{figure}[t]
  \includegraphics[width=\textwidth]{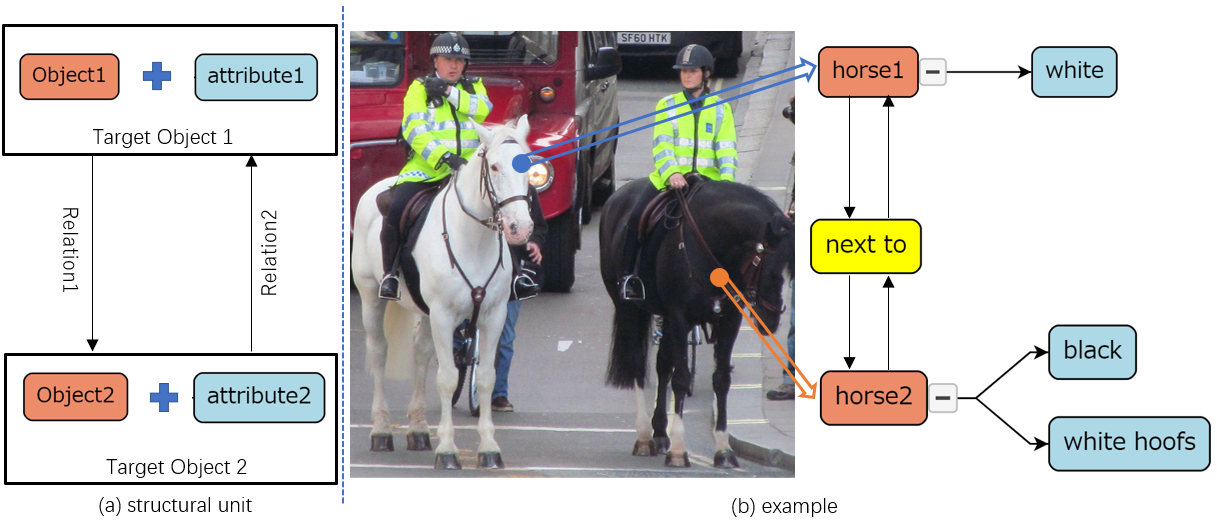}
  \caption{Illustration of the structural unit (a) as an object-oriented representation and (b) its example in this work. Taking two objects as an example, the attributes as objects can be directly manipulated, and attached to the objects with user operations. The relationships can then be constructed among the objects.}
  \label{fig:framework}
\end{figure}

In this work, we propose an interactive drawing user interface, SGDraw for creating scene graphs using object-oriented representation, as shown in Figure \ref{fig:framework}. The proposed system can achieve more comprehensive and detailed results than the conventional annotation interfaces. SGDraw adopts object-oriented representation for scene graph annotation by considering objects, object relationships, and object attributes. SGDraw can take advantage of the user's cognitive ability in the drawing process from target images. We verified the proposed interface in the image dataset and conducted a comparison study between the conventional annotation method using bounding boxes and our proposed approach. The evaluation results demonstrated the superiority of the proposed approach in drawing comprehensive and detailed scene graphs.

We list our main contributions as follows:

\begin{itemize}
    \item We propose an object-oriented annotation approach of scene graphs, which adopts the structural unit of the object, its attributes, and the relationships among objects. In this way, we can describe the image in detail and create a comprehensive and detailed scene graph.
    \item We design a web-based lightweight annotation tool that is easy to deploy and operate so that the proposed SGDraw can draw scene graphs interactively.
    \item We conduct user experiments to verify the performance of our proposed object-oriented annotation approach for annotation, especially the effectiveness of annotation on complex images.
\end{itemize}

\section{Realted Woek}
\subsection{Scene Graph Generation}

 As the expansion of VRD, many methods have been proposed for SGG to detect objects and their relationships~\cite{dai2017detecting,yikangvip,li2017scene,liang2017deep}. Subsequent works significantly improved performance on all-shot recall~\cite{yang2018graph,zellers2018neural,zhang2019graphical} and actively explored the zero-shot generalization problem~\cite{suhail2021energy,tang2020unbiased}. These methods generated scene graphs in two different approaches~\cite{li2018factorizable}. The conventional methods of scene graph generation adopt both object detection and pairwise predicate estimation. The objects are first detected given a bounding box, and then the predicates are predicted using conditional random fields~\cite{dai2017detecting,johnson2015image} or a classification approach~\cite{kolesnikov2019detecting,qi2019attentive}. The other approach is to jointly infer the objects and their relationships based on the suggestions of object regions~\cite{survey}. However, these approaches focus on objects and their relations without attention to object attributes that significantly weigh the image description. We consider that equalizing attributes over objects and their relations can improve the quality of scene graphs.

\subsection{User Interface for Scene Graph}

Visual Genome dataset~\cite{krishna2017visual} provided an annotation tool and used Amazon's Mechanical Turk (AMT) to collect scene graph data~\cite{johnson2015image}. This tool can obtain instances of objects and their relationships by users drawing bounding boxes to identify objects and describing relationships between pairs of objects through text box input. However, this interface has no intuitive, real-time visual output and is difficult to modify. GeneAnnotator~\cite{zhang2021geneannotator} provided a semi-automatic annotation tool for scene graphs, which provides rule-based relationship recommendation algorithms that can reduce the annotation effort. But this tool only focuses on the traffic images and outputs scene graphs without attributes and suitable layout. An interactive interface was presented for scene graph drawing~\cite{zhang2022interactive}, which recommended attributes for users to reduce the drawing time cost. However, this work lacked image visualization and extensive evaluations of the proposed functions. In this work, we aim to provide an interactive drawing interface for scene graphs by the object-oriented method.  

\subsection{Object-Oriented Interface}

Object-oriented methods have been widely used in design tasks. An object-oriented interface~\cite{popoola2021object} was proposed to generate and manipulate polyhedral data flow graphs. An integrated set of object-oriented pixel-based vectorization algorithms~\cite{song2002object} was then presented for various classes of graphic objects in engineering drawings. These interfaces performed interactions that were previously tedious or even impossible with a coherent and consistent interaction experience throughout the entire interface. Next, an object-oriented drawing approach~\cite{xia2016object} was proposed to represent attributes as objects that can be manipulated directly. Considering the basic components of the scene graph, the objects, object attributes, and their relationships could be manipulated directly by the user in this work.

\section{Scene Graph Drawing}
In this work, we aim to propose a drawing interface for scene graph annotation and creation. The basic structural unit of the proposed approach is represented in Figure~\ref{fig:framework}. 

\subsection{Scene Graph}
Since the location relationships between objects, rather than coordinate information, are more used in most of the follow-up research. In this work, scene graph creation is achieved by objects without bounding boxes, and we provide the open-ended and free-form manipulation of scene graph nodes. The scene graph from an image is defined as follows by a tuple of $(O',R)$ with   object attributes $O'$ and relations $R$:

\begin{equation}
    P(G|I) = P(O',R|I),
\end{equation}

\noindent
where $P$ stands for probability, $I$ represents the input image, and $G$ is the desired target scene graph containing object attributes and relations. Specifically, the image is partitioned into a set of objects $O'=\{o'_1,...,o'_n\}$ corresponding to each of the $n$ objects with attributes in the image. $R=\{r_1,...,r_l\}$ is the set of relations between pairs of objects.

\begin{equation}
    P(o') = P(o,A),
\end{equation}

\noindent
where $o'$ is the target object with attributes to be generated, containing the input target object $o$ and its attribute set $A$. Each object comes with its unique set of attributes $A=\{a_1,...,a_m\}$ corresponding to the $m$ attributes of the target object. 

In this work, we handle the object and its attributes as a manipulable object unit. By linking the pairwise relationships between different object units, we can perform the task of drawing the original image to the scene graph. In contrast to the conventional scene graph annotation task that focuses on the objects and their relationships, we also focus on the object attributes that are easily ignored and that account for a large proportion of the images. The drawn scene graph can be more consistent with the image content and represent the image scene in more detail.

\subsection{Object-Oriented Representation}

\begin{figure}[t]
  \includegraphics[width=\textwidth]{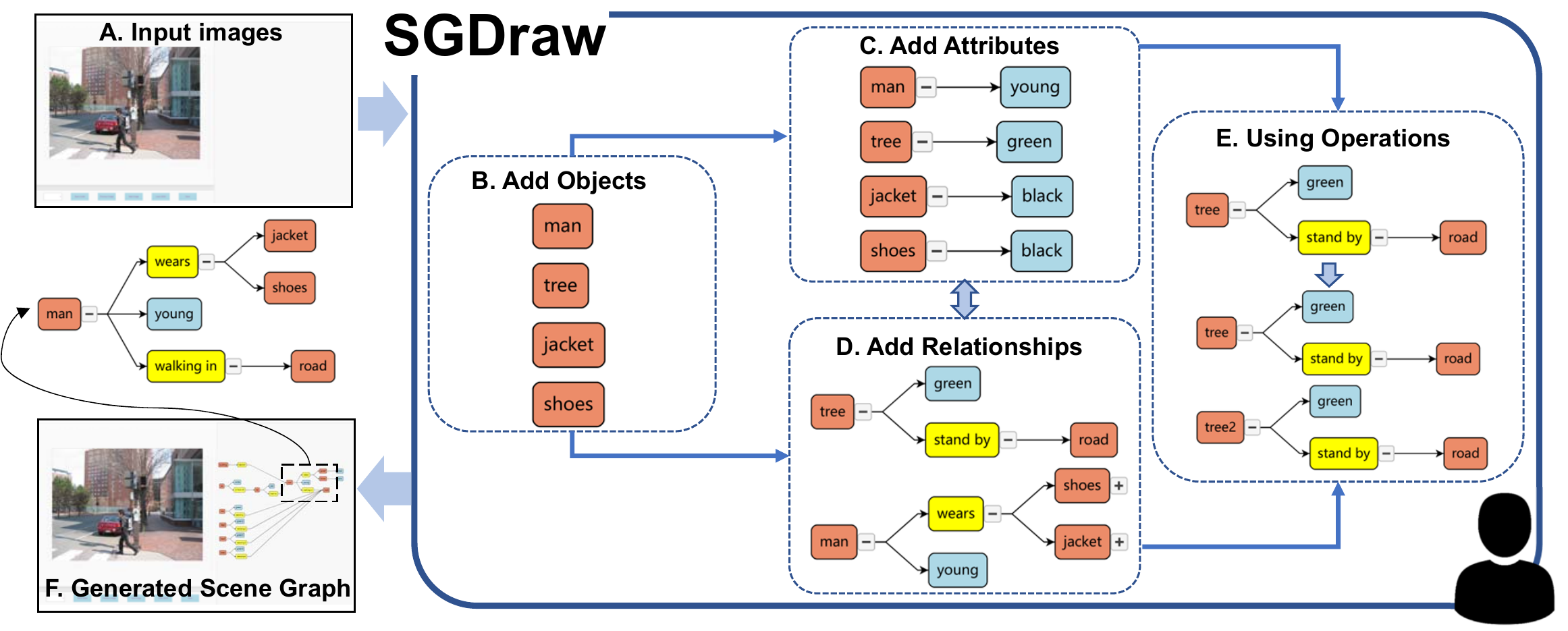}
  \caption{Workflow of the proposed interface. (A) With a task image inputted to the SGDraw, the user starts (B) adding objects, and then they can (C) add attributes or (D) relationships. After that, they can choose to (E) use the auxiliary operations (like cloning, as shown in the figure) to change the structure of the graph. Repeat these steps until (F) obtains the desired scene graph.}
  \label{fig:ui}
\end{figure}

Inspired by the object-oriented drawing approach~\cite{xia2016object}, we aim to propose an object-oriented representation of scene graphs for image annotation and graph editing tasks. In this representation, we can extend attributes to objects directly for manipulation, as shown in Figure \ref{fig:ui}. Specifically, the proposed object-oriented approach has the following features:

\noindent
\textbf{Abstraction.} The input images may contain repetitive contents, such as many horses and trees (Figure \ref{fig:framework}, Figure \ref{fig:ui}(A)). For repetitive objects, concrete objects may contain similar attributes. Therefore, we can abstract them into a class, such as the class of horse. Through the process of class abstraction, we can reuse the scene graph data to enhance the versatility and scalability of the proposed system.

\noindent
\textbf{Uniqueness.} As shown in Figure \ref{fig:ui}(B), each object has its unique identifier by which the corresponding object can be located. The identity of the annotated objects will not change during the whole task. Different objects in the same category are given different identifiers, such as ``horse1" and ``horse2" in Figure \ref{fig:framework}. By assigning the identifiers to objects (Figure \ref{fig:ui}(D)), we can correspond scene graph nodes to unique objects in the image to create detailed and higher-quality scene graphs.

\noindent
\textbf{Polymorphism.} Polymorphism refers to the fact that objects have diverse attributes.  Multiple different attributes can be assigned to the same object simultaneously, and the same attribute can be assigned to multiple different objects simultaneously. For example, in Figure \ref{fig:framework}, in the case of several people and horses, different attributes like black color for fur and white color for hoofs can be assigned to ``horse2". In addition, as shown in Figure \ref{fig:ui}(C), attributes like black can be assigned to the ``jacket" and the ``shoes". The diversity of attributes allows each object to correspond to the same or different attributes, which enhances the flexibility and reusability of the system.

\begin{figure}[t]
  \includegraphics[width=\textwidth]{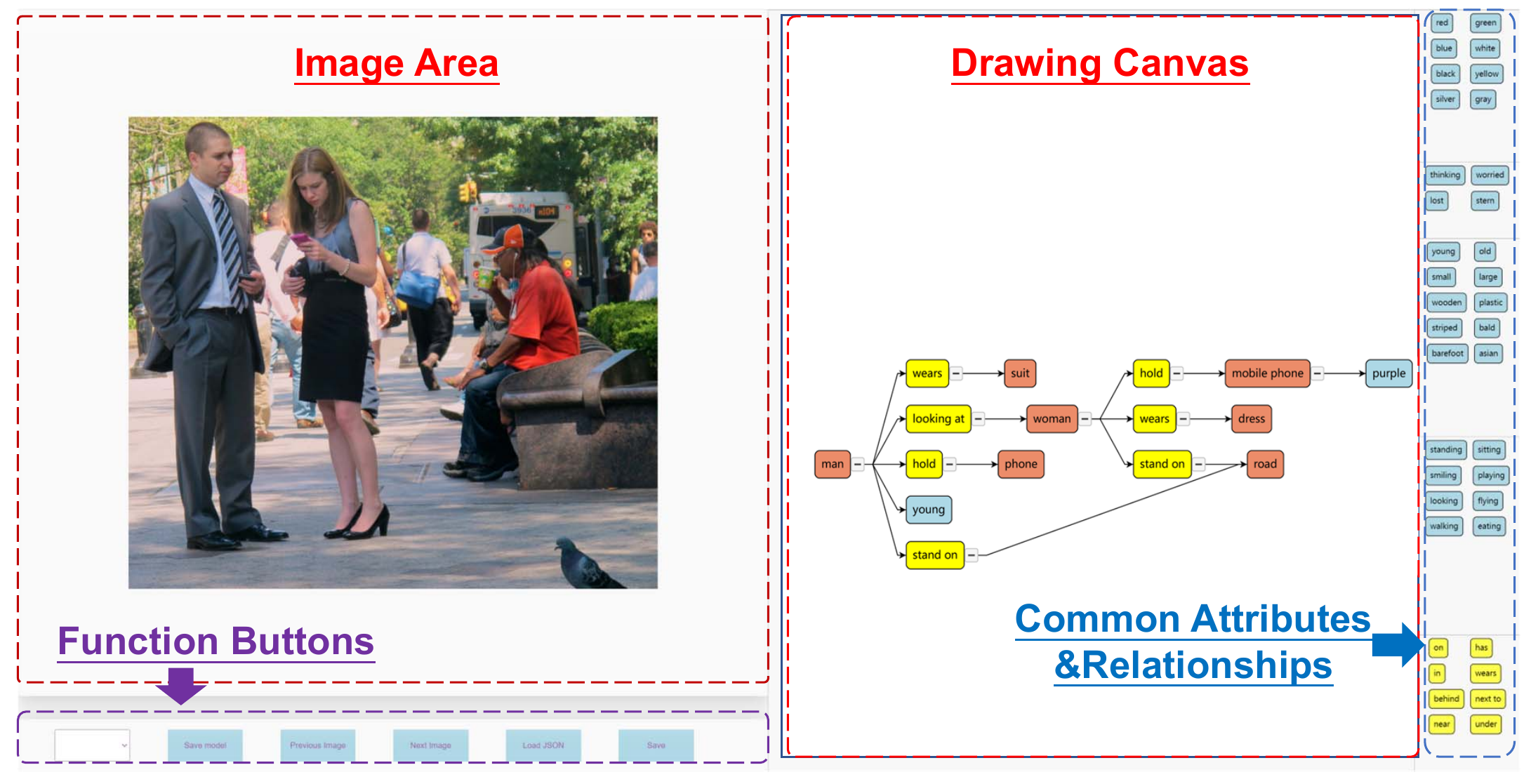}
  \caption{Screenshot of the SGDraw interface. On the left is the input image display area, and on the right is the scene graph operation and real-time generation area. Meanwhile, on the right side of the interactive interface, we provide some common attributes and relationships for users to choose and reference.}
  \label{fig:uiface}
\end{figure}

\noindent
\textbf{Inheritance.} Inheritance refers to the ability to define and implement objects on the basis of objects that already exist. The attributes and relationships associated with existing objects can be used as their own content, and new attributes and relationships can be added. Specifically, the user can select objects to store after drawing the scene graph. As shown in Figure \ref{fig:uiface}, all attributes and relationships of the selected object can be stored in the drop-down box in the left corner of the interactive interface. The user can select the object at any time to inherit all previously stored attributes and relationships in subsequent operations (Figure \ref{fig:ui}(E)). The use of inheritance provides a hierarchical structure between objects, attributes, and relationships. Inheritance makes it possible to share common features and increase the reusability of the system.

\subsection{SGDraw Interface}

In order to explore the effectiveness of the proposed SGDraw, we developed a web-based user interface, as shown in Figure~\ref{fig:uiface}. The objects, attributes, and pairwise relationships between objects are considered as objects that can be manipulated directly. We provide an open and fully accessible annotation tool that takes full advantage of the user's cognitive ability to perform the task of annotating any kind of image and creating their corresponding scene graphs in real time. In contrast to the previous approach that requires drawing the bounding box on the image first and then describing the objects inside the box, SGDraw can help users easily add/delete and drag nodes depending on their knowledge of the image content until a satisfactory scene graph is drawn. Specifically, we provide the SGDraw with the following interaction features:

\begin{figure}[t]
  \includegraphics[width=\textwidth]{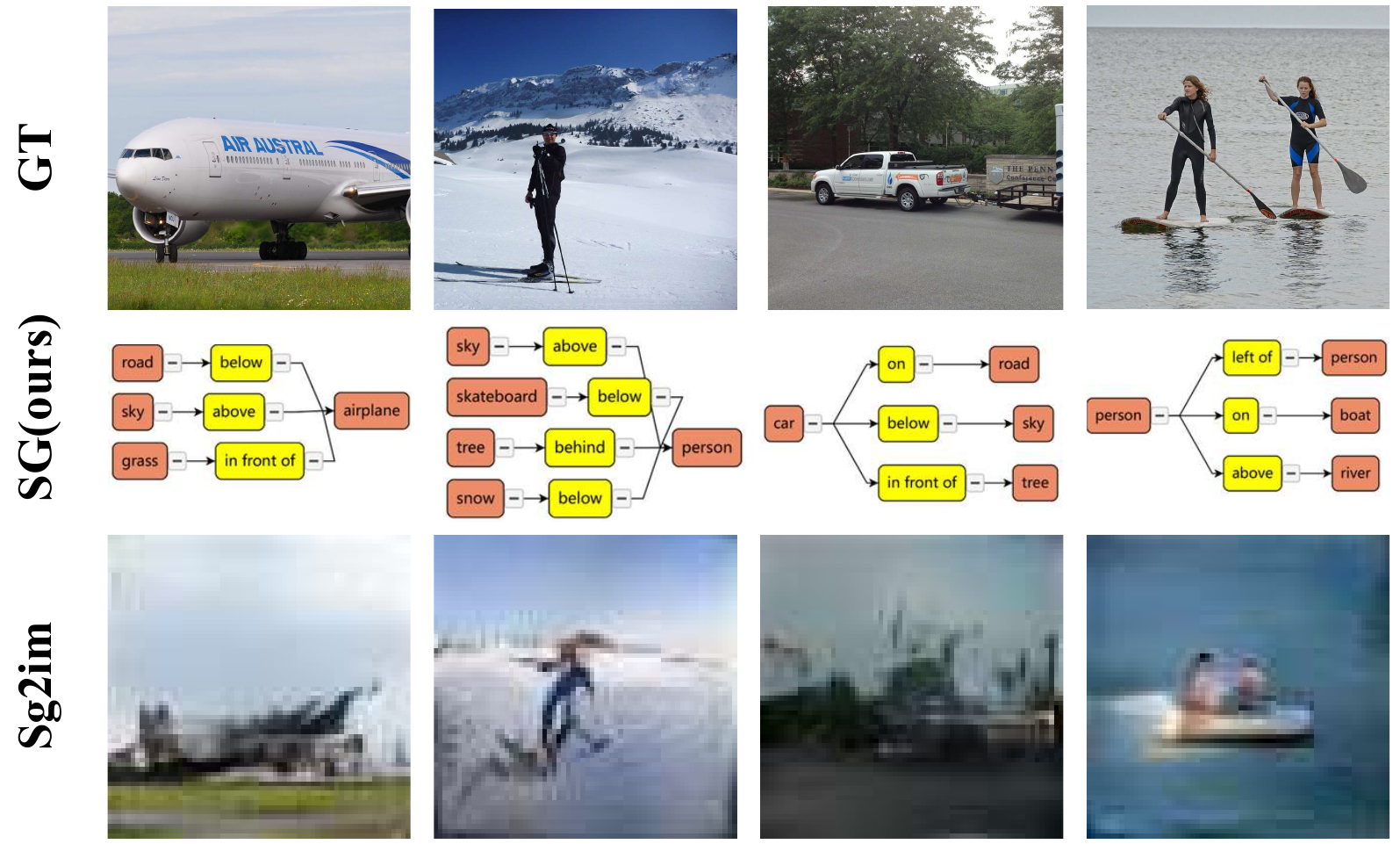}
  \caption{Example of image generation tasks with the drawn scene graphs with ground truth images from the VG dataset (1st row), the drawn scene graphs (2nd row), and the generated images (3rd row).}
  \label{fig:sg2im}
\end{figure}

\noindent
\textbf{Direct operation of the scene graph.} In previous interfaces~\cite{johnson2015image,zhang2021geneannotator}, the users cannot observe the visualized scene graphs in real time and modify the input content. To solve this issue, we consider the input of the scene graph as a drawing. We provide three kinds of nodes: red nodes for objects; yellow nodes for relationships; blue nodes for attributes. The user can add a new object and attribute by simply right-clicking on the background and object. A relationship node can be generated when clicking two different objects. We also provided common attributes and relationships for reference, which were derived from the common ranking in the scene dataset~\cite{xu2017scene}. The common part is not displayed on the screen initially to avoid giving hints to the user and making the drawn scene graph monotonous. It will be shown by pressing the key when the user needs it and will be automatically closed after use. We believe that common attributes and relationships can simplify users' input and facilitate creative thinking.

\noindent
\textbf{Various auxiliary operations.} We designed the auxiliary operations to modify the graph structure and simplify the input process, including removing, cloning, undoing, zooming, collapsing, and dragging. We implemented the undo function to reduce the loss caused by user error operations, and the removing function to help users change the graph structure freely. We implemented the zooming and collapsing functions for complex scene graph drawing to help users understand the overall graph structure. The dragging function allows users to optimize the layout to achieve visual results that are easier to operate and observe. We also added the cloning function for similar attributes and relationships to increase the efficiency of the drawing process. These operations are directly conducted on the scene graphs with real-time feedback to help users clearly understand the changes in the scene graphs.

\noindent
\textbf{Flexible data archive.} SGDraw provides two ways to save the scene graph: JSON files to facilitate subsequent vision tasks, and SVG files so that the visual scene graphs become available to study and understand the scene images more intuitively. For JSON files, as shown in Figure~\ref{fig:sg2im}, the created scene graph results can be adapted to various vision tasks, such as image generation. In particular, we suggested the user write objects and relationships due to the limitations of the generation algorithm (sg2im~\cite{johnson2018image}). For SVG files, users can draw the images directly and make the image interactive by changing part of the code, and then inserting it into HTML while viewing through the browser. SGDraw allows the users to load the graph and modify it for further usage, which can improve the reusability and maintainability of the scene graph data.

\section{User study}

In the user study, we confirmed SGDraw for scene graph drawing and collected feedback from potential users about the effectiveness and usability of the interface. We asked 14 participants (aged 25 to 30, eight males and six females) who are graduate students with knowledge of machine learning. Especially, we designed two experiments: a comparison study and a user experience study. We reproduced the previous interface for scene graphs~\cite{johnson2015image} in the comparison study, and the task images used in the experiments were selected from the Visual Genome dataset~\cite{krishna2017visual}. The evaluation process consisted of three stages:

\noindent
\textbf{1) Introduction and Training (10–15 minutes).} We first introduced the background of the scene graphs. The experimenter then guided the participants to explore the functions and workflow of the previous interface and SGDraw, and we intentionally provided a simple image for participants as an example. This was done for two reasons: First, it ensured that participants had a full understanding of the interfaces, and second, it allowed participants the opportunity to explore and demonstrate understanding.

\noindent
\textbf{2) Experience and Usage (20–30 minutes).} To avoid the proficiency effect caused by the interface order for usage, participants were divided into two groups. One group used the previous interface first and then used SGDraw. The other group used SGDraw first and then used the previous interface. Each participant was asked to draw the scene graphs in two interfaces based on two different images. The system automatically counted time without user knowledge. When the participants thought they had finished describing the content of the images, the experimenter collected the results. 

We obtained seven task results from the experimenters finally. Each task included four results for an image, two results from the SGDraw, and two results from the previous interface. We collected the objects, attributes, and relationships that participants described on the scene graphs. After collecting the data, we calculated the instances (including objects, attributes, and relationships) per minute for each interface to compare the efficiency of the two interfaces.

\noindent
\textbf{3) Questionnaire and Interview (10–15 minutes).} The participants next completed questionnaires about the system. The questionnaires were composed of a 5-point Likert scale (1 for strongly disagree and 5 for strongly agree) to collect the participants' experience of SGDraw. The questions on the first questionnaire, based on the System Usability Scale (SUS), were used for the usability measurement of an interface. The second questionnaire collected usability feedback on every function. The interview then consisted of open-ended questions that were asked to gain the users’ feedback on workflow, utility, and usability.

\section{Results}

\subsection{Drawn Scene Graph}

\begin{figure}[t]
  \includegraphics[width=\textwidth]{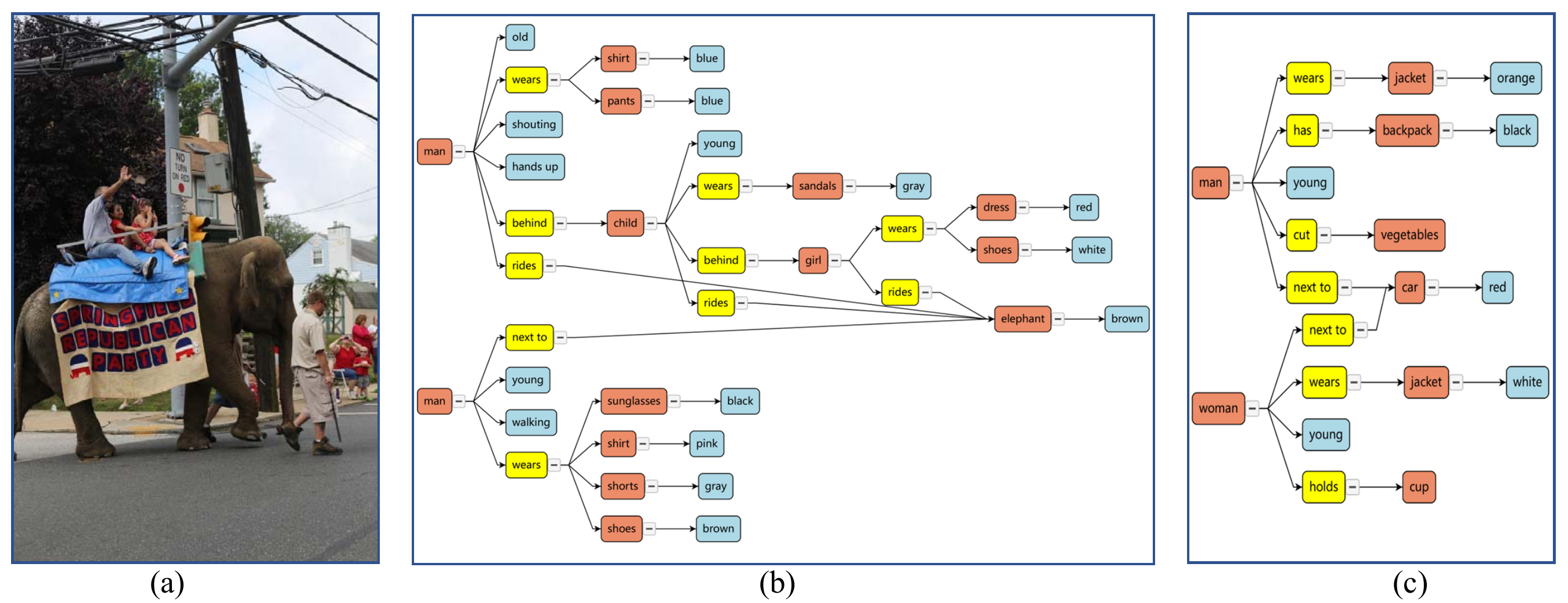}
  \caption{Examples of scene graphs completed by SGDraw with (a) an image input from the VG dataset, and (b) a completed scene graph. (c) shows the scene graph freely designed from the text input ``a couples drive to the forest to have a picnic''.}
  \label{ui_result}
\end{figure}

Figure~\ref{ui_result} shows that SGDraw could help users complete the scene graph drawing task. Without the given ground truth image, the users would use SGDraw to express the image they desired. Users will not be limited to using the scene graphs in the dataset for subsequent tasks such as generation and retrieval. In addition, compared with the results of previous research that did not visualize ~\cite{johnson2015image} or whose layout was messy~\cite{zhang2021geneannotator}, the drawn scene graphs in SGDraw are constructed based on an automatic tree layout, which enhances the readability and aesthetic of the users' results. The automatic tree layout helps but does not limit users. The results will be expanded automatically and hierarchically, and users are allowed to change the graph structure with any node or link. In the results, red nodes represent objects, blue nodes represent attributes, and yellow nodes represent relationships. The nodes classified by colors strengthen the user's understanding of specific details and further enhance the comprehensibility of the results on top of the clear layout. The joint storage of the visual scene graphs and the textual results are more conducive to the user's subsequent use and modification.

\subsection{Comparison Study}

\begin{figure}[t]
  \includegraphics[width=\textwidth]{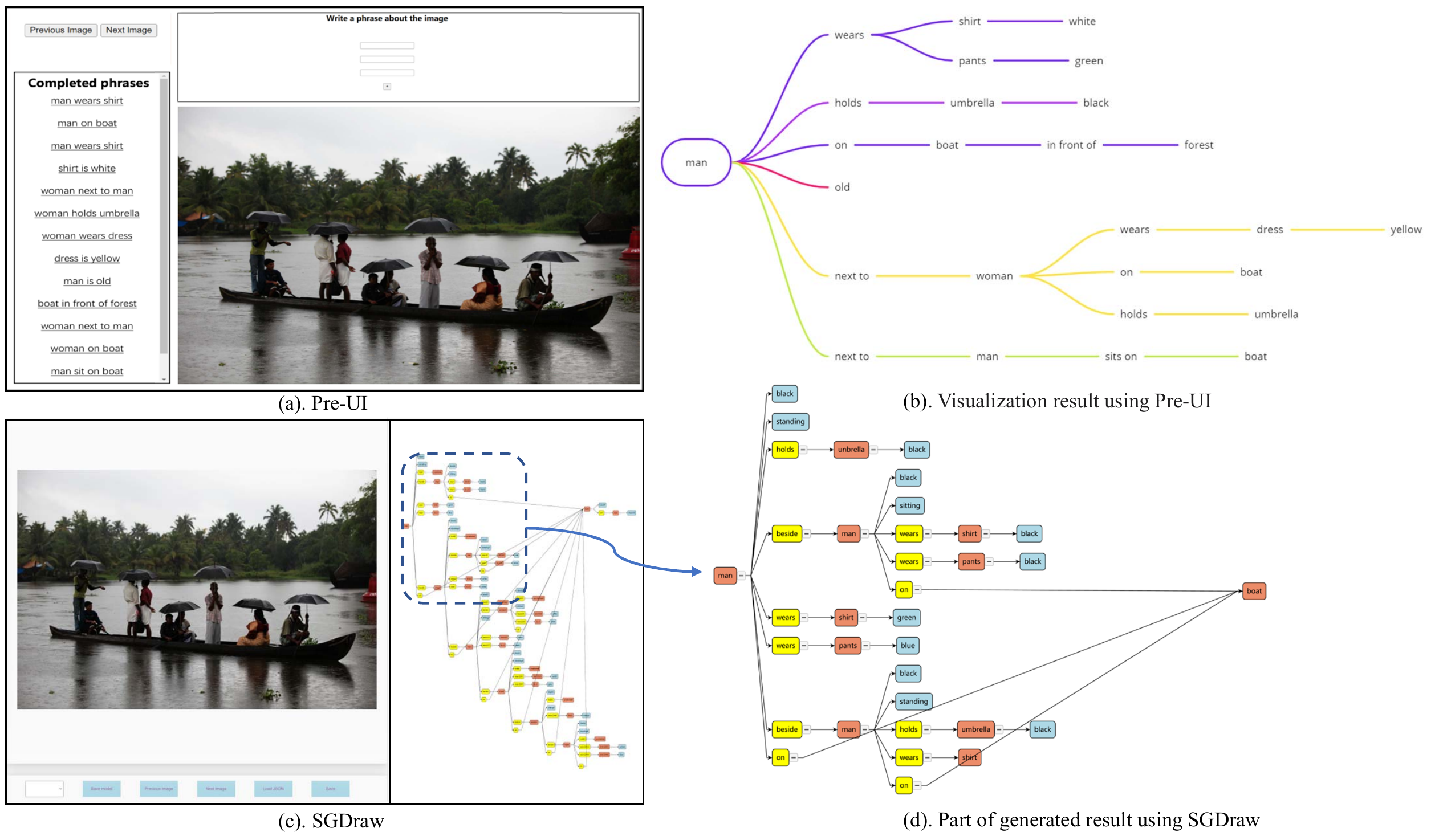}
  \caption{We conducted the comparison study on Pre-UI~\cite{johnson2015image} and the propsed SGDraw. (a) shows the Pre-UI that we reproduced, and (b) shows the visualization result of the graph. (c) is the screenshot of SGDraw, and (d) is the drawn result.}
  \label{fig:compare_ui_result}
\end{figure}

\begin{figure}[t]
\centering

\subfigure[]{
 \label{fig:compare_time} 
 \includegraphics[width=0.45\textwidth]{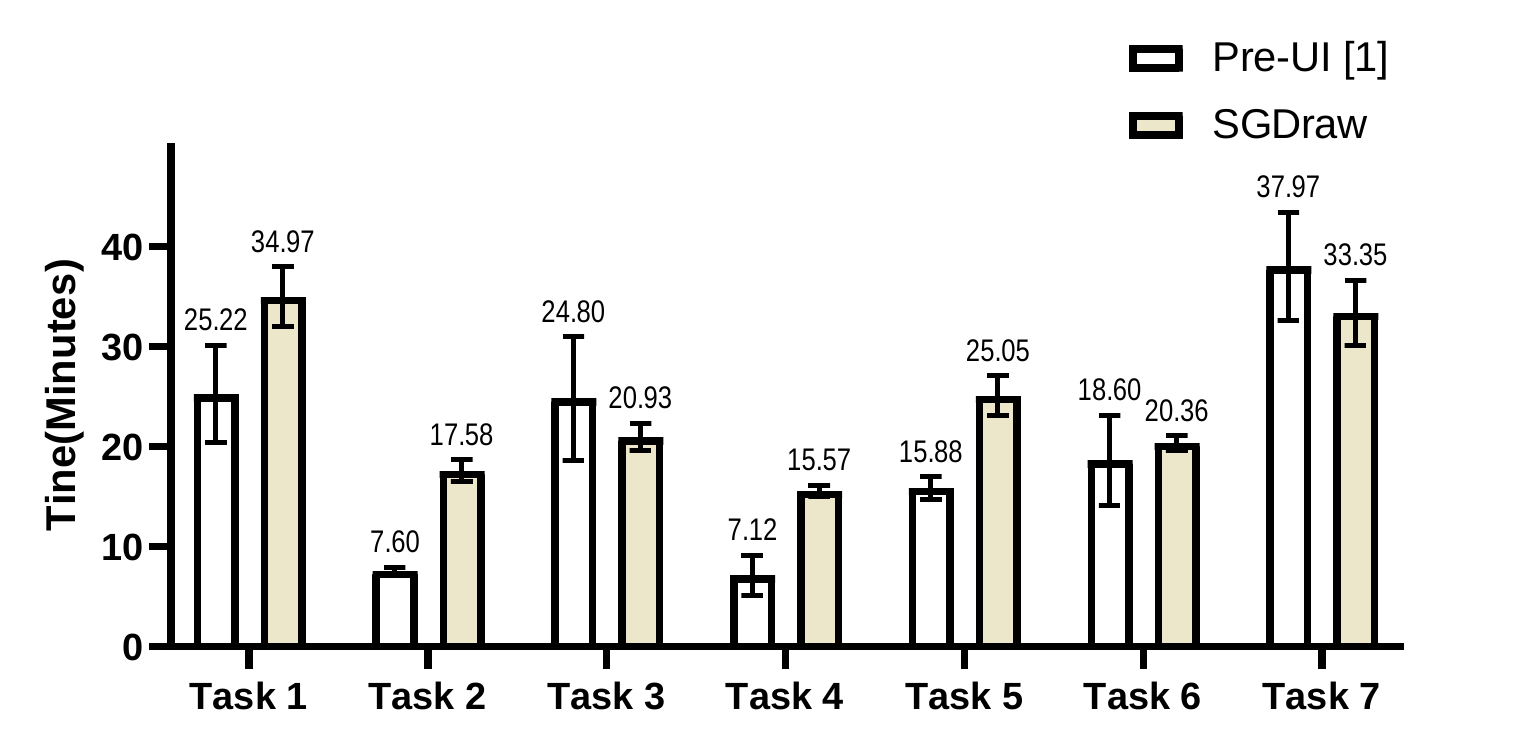}
}
\subfigure[]{
\includegraphics[width=0.45\textwidth]{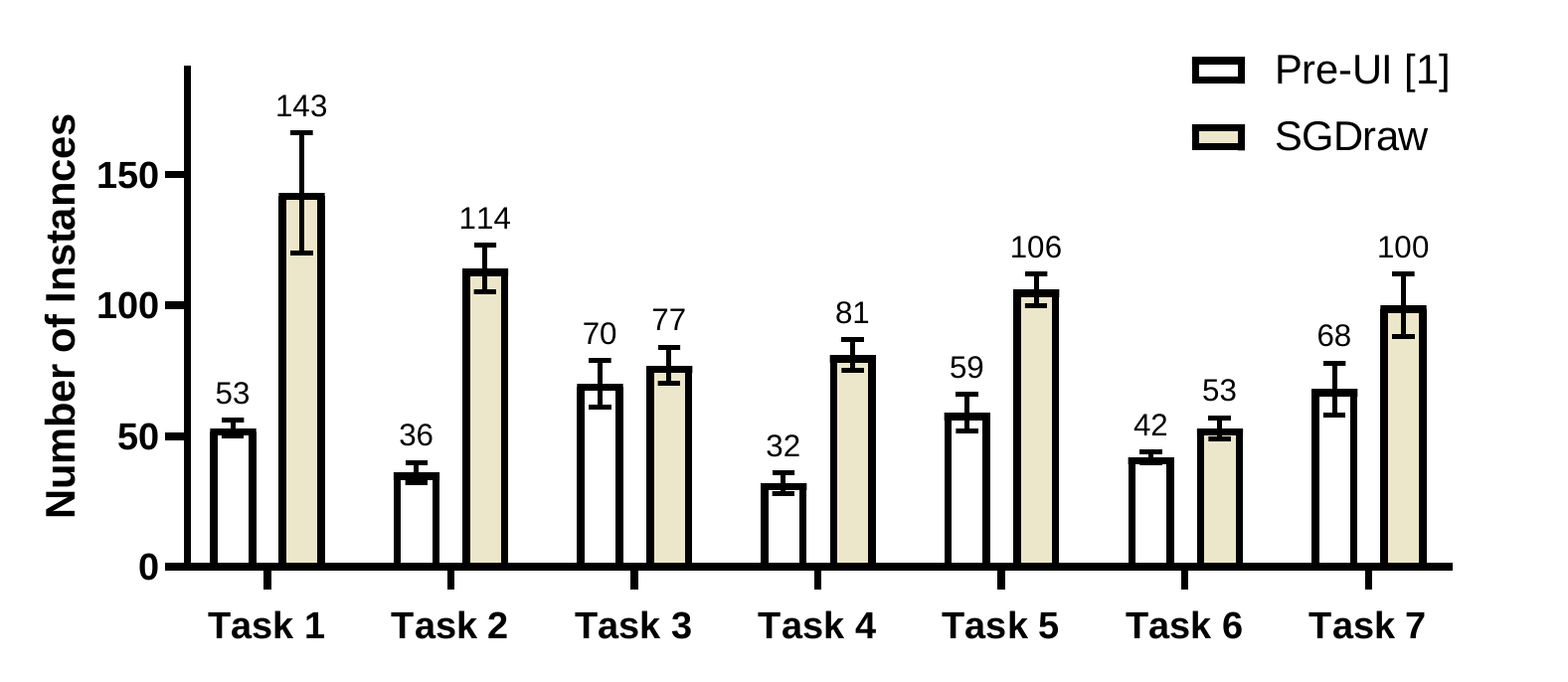}
\label{fig:compare_instance}
}

\caption{(a) The time cost of tasks. The system counts time automatically and implicitly. (b) For each task, experimenters calculated the instances (including objects, attributes, and relationships) in two interfaces. The number of instances that are completed by the SGDraw is more than pre-UI~\cite{johnson2015image}.}
\label{fig:}
\end{figure}

In the comparison study, participants are asked to draw scene graphs with the previous interface (pre-UI) and SGDraw respectively, with two random images, as shown in Figure~\ref{fig:compare_ui_result}. We recorded the number of objects, attributes, relationships, and time cost.  As shown in Figure~\ref{fig:compare_time}, SGDraw does not achieve a time advantage on all drawings, but it does achieve the advantage of instance quantity (including objects, attributes, and relationships.) as shown in Figure~\ref{fig:compare_instance}. To verify efficiency, we calculated the number of instances that could be completed for each task per minute. In all experiments, SGDraw achieved the advantage of efficiency. 

\begin{figure}[t]
  \includegraphics[width=\textwidth]{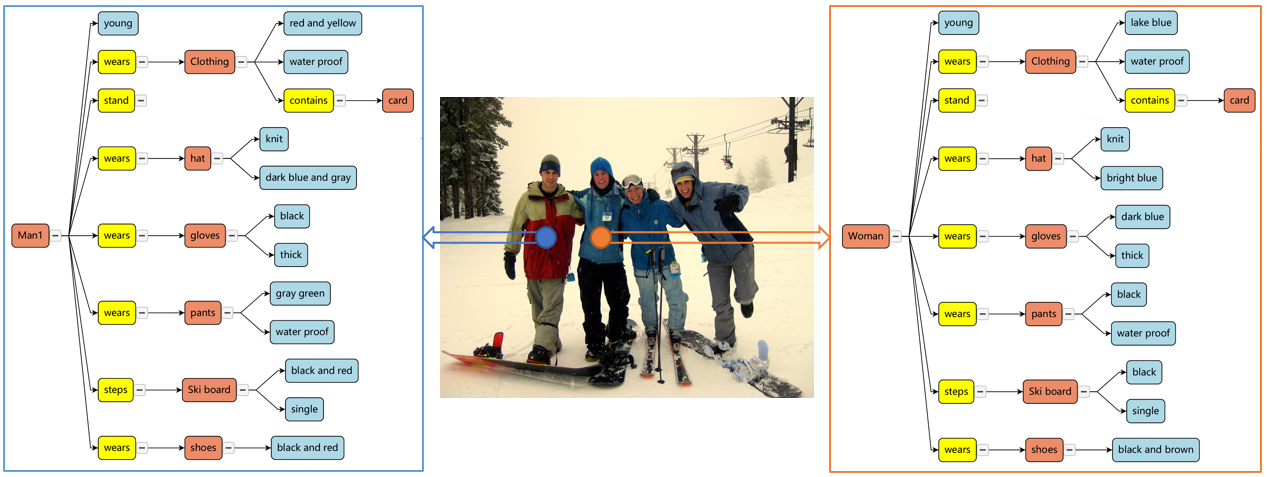}
  \caption{Scene graph drawing results from the user study. One participant used SGDraw to complete the complex scene graph for the task image by the auxiliary tool. The similar attributes and relationships of objects in the image made our interface facilitate the scene graph drawing process.}
  \label{fig:result}
\end{figure}

For Task 1, SGDraw had a larger time cost but better performance in instances quantity (4.9 instances per minute, and 2.1 instances per minute for pre-UI). As shown in Figure~\ref{fig:result},  the objects of the task have the same attributes and relationships. After completing the scene graph of one person, the participants only needed to use the cloning function with a few modifications using SGDraw to complete the whole task. For Task 7, SGDraw achieved higher scores in both time cost and instance quantity than pre-UI. After the interview and analysis, we found that the recommendation part played a significant role. For complex images with blurry objects, participants had to think about and describe the scene for a long time. The common part helped participants facilitate the input by the given attributes and relationships. In addition, the common part better cued them about attribute and relationship categories that could be considered when participants were out of ideas. For example, the ``wooden" given in the common part, users can quickly associate ``metallic" and ``plastic", even though these attributes are not commonly used compared to color gestures. This part enlightened the participants' minds and reduced their thinking pressure. Task 6 had the smallest difference in the efficiency of the two interfaces. In this task, the image composition was simple and clear; thus, the participants understood the main objects explicitly in the image. Based on less similar attributes and relationships, the scene graph was completely dependent on the participants to finish it, so SGDraw achieved fair efficiency for this task.

\subsection{User Experience Study}

\begin{figure}[t]
  \includegraphics[width=\linewidth]{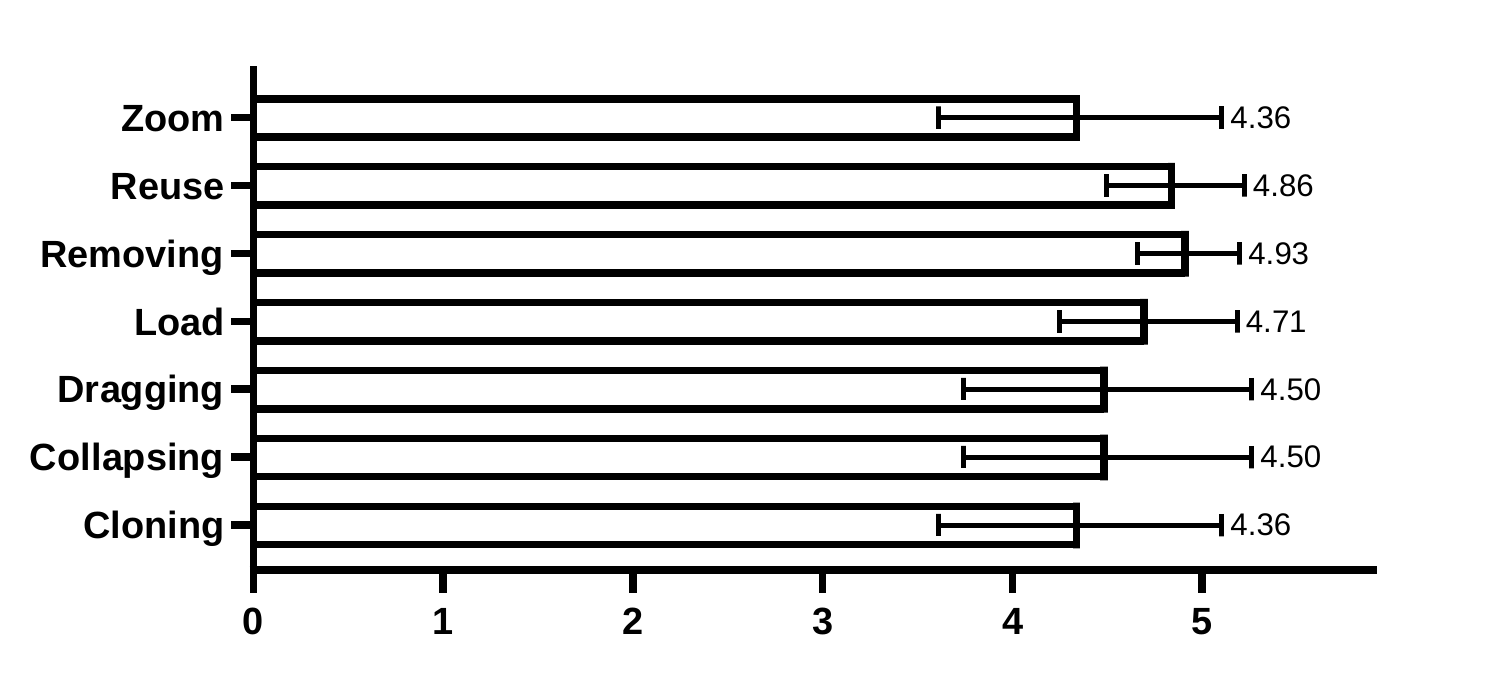}
  \caption{Result of five-point Likert-scale responses to “Do you agree that function is useful for your scene graph drawing tasks?”}
  \label{fig:function}
\end{figure}

%\begin{figure}[t]
  %\includegraphics[width=\linewidth]{img/sus_revised_5.pdf}
  %\caption{The result of the SUS questionnaire.}
 % \label{fig:SUS}
%\end{figure}

\noindent
\textbf{Workflow.} Participants were interviewed about how the SGDraw integrated with their workflow. Almost all responded positively; the specific feedback from the participants included the following: \textit{``After learning the simple operations, my workflow in this interface was fluid."} (P2). \textit{``In traditional painting, the overall composition of the picture is firstly carried out to determine the position of the objects. In addition, some people have the habit of proceeding to the next object after describing one object. Either of these two workflows can be well implemented in this interface, and both have specific functional assistance."} (P5). Participants also pointed out that there is some disadvantage to the SGDraw. \textit{``The system is smooth and can complete tasks well, but a lot of use of the right click will be unfriendly to people who are used to the left click."} (P10).

According to the interviews, SGDraw obtained a good performance in the workflow. The participants agreed that SGDraw is smooth and helpful for users with different backgrounds. We also found that few function operations may cause discomfort for users with specific computer habits.

\noindent
\textbf{Utility.} Participants were also asked to rate the usefulness of each of the functions. The functions of SGDraw received a high agreement, as shown in Figure~\ref{fig:function}. All of the functions obtained more than four points, which verified that the proposed functions could achieve simple operation and good user experience. Among the various functions, SGDraw's ability to delete the link and node with their subtrees was strongly praised by participants. In addition, the scores of some functions were underrated due to the limitations of usage fields. For example, in scenes with complex and repetitive objects (such as stadiums, and parking), the cloning function always brought a good experience to users. However, in other simple scenes, cloning was not so important.

Based on the various functions, SGDraw could save and reuse scene graphs was strongly favored by participants. Users could thus build their datasets in turn to facilitate later operations. Some specific feedback included the following: \textit{``I like the Cloning function; it frees me from repetitive typing work. Being able to replicate attributes and relationships together is simply awesome."} (P5). \textit{``The Removing function combines delete and rollback, which is a sensible approach. I can easily do what I want, and reduce the cost of my mistakes."} (P7). The functions in SGDraw have high approval which indicates that the various functions are valued and desired. Users can achieve the desired results through the functions we provided in SGDraw, which makes the input process more convenient and quick.

\begin{table}[t]
\caption{The result of the SUS questionnaire. $\Uparrow$ indicates that higher scores are better; $\Downarrow$ for the other case.} 
\label{table:sus}
\begin{center}     
%\resizebox{\textwidth}{
\begin{tabular}{|c|l|c|}
\hline
 & Questions & Mean  \\
\hline
    1 & \makecell[l]{I would like to use this system frequently. $\Uparrow$} & 3.86\\ \hline
    2 & \makecell[l]{I found this system unnecessarily complex. $\Downarrow$}  & 2.21\\ \hline
    3 & \makecell[l]{This system was easy to use. $\Uparrow$}  & 4.36\\ \hline
    4 & \makecell[l]{I would need the support of a technical person to be able to use\\ this system. $\Downarrow$} & 2.14 \\ \hline
    5 & \makecell[l]{I found the various functions in this system were well integrated. $\Uparrow$} & 4.21\\ \hline
    6 & \makecell[l]{I thought there was too much inconsistency in this system. $\Downarrow$} & 1.29\\ \hline
    7 & \makecell[l]{I would imagine that most people would learn to use this system\\ very quickly. $\Uparrow$} & 4.57\\ \hline
    8 & \makecell[l]{I found this system very cumbersome to use. $\Downarrow$} & 2.00\\ \hline
    9 & \makecell[l]{I felt very confident in using this system. $\Uparrow$} & 4.14\\ \hline
    10 & \makecell[l]{I needed to learn a lot of things before I could get going with\\ this system. $\Downarrow$} & 1.57\\ \hline
\end{tabular}
%}
\end{center}
\end{table}

\noindent
\textbf{Usability.} 
Participants were asked to complete questionnaires about interface usability. The questionnaire was designed with the question items using SUS, which provides an overall usability assessment measure consisting of 10 items. We use a five-point Likert scale ( one for strongly disagree and five for strongly agree).

As shown in Table~\ref{table:sus}, SGDraw achieved good performance with high scores in positive items and low scores in negative items. Almost all participants agreed that SGDraw is easy to use (Scored 4.36) and that most people learned to use it quickly (Scored 4.57). They strongly disagreed that there were too many inconsistencies in SGDraw (Scored 1.29) and many things to learn before usage (Scored 1.57). The total SUS availability score of SGDraw was 78.9 out of 100. Based on the results of previous research~\cite{bangor2008empirical}, this total score shows that SGDraw should be judged to be acceptable, and the Adjective Rating should be ``excellent."

All participants commented that the interface is intuitive and easy to learn. The participants noted the following: \textit{``The common attributes part is great! It helps me spread my mind when I don't know what to write, and it's usually hidden without making me rely on it."} (P11). \textit{``No need for more knowledge, it seems friendly for most people."} (P1). \textit{``It is interesting to operate directly on the results. You can intuitively see the changes in the graphs. Compared with the textual results, the graphs are always easier to understand."} (P4). Several participants pointed out that more functions are needed, such as selecting the common attributes by themselves: \textit{``Quite smart and consistent functions it has created. For people with different tasks, the common attributes required are also different. Perhaps it is a better way to let users set the common attributes themselves."} (P6).

Based on the interviews, the visualization results and direct operations on the graphs have received favorable reviews. We noted that the common area on the right side of SGDraw sparked discussion. Participants agreed that it can simplify operations and inspire users' minds, but it should also be able to be smarter. At the moment, we only display some common attributes fixedly. However, different users need different common attributes. Allowing users to change the common area by themselves will improve the interface's interactivity. 

\section{Conclusion}

This work presented SGDraw, a scene graph drawing interface based on the object-oriented method, to help users draw scene graphs succinctly and conveniently. We designed a set of complete functions and presented common tools to ease user operations. Finally, we conducted a user study to verify our proposed interface. We found that the proposed interface can especially work well with complex scene graphs. 

For limitations of this work, the frequent right clicks may make users who are accustomed to left clicks uncomfortable. In addition, the function design can be improved to help users obtain free and better experiences. As future work, we envision combining SGDraw with the scene graph generation algorithms, which can help users semi-automatically complete the scene graph drawing. Beside of scene graph, we also plan to apply the object-oriented method in other domains, such as illustration drawing~\cite{huang2022dualface} and layout design~\cite{weng2023dualslide}. To facilitate the subsequent work using SGDraw, we released the source code of this work in an open source platform.  

\section*{ACKNOWLEDGEMENTS}
We thank all the participants in our user study. This work was supported by JAIST Research Grant, and JSPS KAKENHI JP20K19845, Japan.
\bibliographystyle{splncs04}
\bibliography{refs}

\end{document}